\pdfoutput=1

\documentclass[11pt]{article}

\usepackage[]{ACL2023}

\usepackage{times}
\usepackage{latexsym}
\usepackage{graphicx}

\usepackage[T1]{fontenc}

\usepackage[utf8]{inputenc}

\usepackage{microtype}

\usepackage{inconsolata}
\usepackage{amsfonts}

\title{ Effectiveness of Data Augmentation for Parameter Efficient Tuning with Limited Data }

\author{Stephen Obadinma$^a$ , Hongyu Guo$^{b,c}$, Xiaodan Zhu$^{a}$ \\
  $^a$Department of Electrical and Computer Engineering, Queen's University, \\ $^b$Digital Technologies Research Centre, National Research Council Canada, \\ $^c$School of Electrical Engineering and Computer Science, University of Ottawa\\ $^a$\ \texttt{\{16sco, xiaodan.zhu\}@queensu.ca} \\ $^{b,c}$ \texttt{\{Hongyu.Guo\}@nrc-cnrc.gc.ca }}

\begin{document}
\maketitle
\begin{abstract}
Recent work has demonstrated that using parameter efficient tuning techniques such as prefix tuning (or P-tuning) on pretrained language models 
can yield performance that is comparable or superior to fine-tuning  while dramatically reducing trainable parameters.
Nevertheless, the effectiveness of such methods under the context of data augmentation, a common strategy to improve learning under  low
data regimes, has not been fully explored. 
In this paper, we examine the effectiveness of several popular task-agnostic data augmentation techniques, i.e., EDA, Back Translation, and Mixup, when 
 using two general parameter efficient tuning methods, P-tuning v2 and LoRA, under data scarcity.  
We  show that  data augmentation can be used to boost the performance of P-tuning and LoRA models, but  the effectiveness of each technique varies and certain methods can lead to a notable degradation in performance, particularly when using larger models and on harder tasks. 
We further analyze the sentence representations of P-tuning compared to fine-tuning to help understand the above behaviour, and reveal how P-tuning generally presents a more limited ability to separate the sentence embeddings from different classes of augmented data.
In addition, it displays poorer performance on heavily altered data. However, we demonstrate that by adding a simple contrastive loss function it can help mitigate such issues for prefix tuning, resulting in sizable improvements to augmented data performance. 


\end{abstract}

\section{Introduction}
While large pretrained language models have 
achieved superior performance and widespread adoption across many NLP tasks~\citep{NEURIPS2020_c8512d14,  bengioDeepLearning}, they often contain hundreds of millions or even hundreds of billions of parameters, which significantly limits their application to tasks in which computation and storage resources are constrained. 

To address this issue, an entire family of techniques called parameter efficient tuning (PET) methods have been developed. Most notably, deep prompt tuning (i.e., prefix tuning or P-tuning) ~\citep{li2021prefixtuning, https://doi.org/10.48550/arxiv.2104.06599, ptuningv2} 
has attracted extensive attention, which, compared to fine-tuning, only tunes trainable continuous embeddings, resulting in a tiny percentage of tuned parameters for each task. Consequently, a single language model can be used for multiple tasks by swapping out the trained prompts on a task basis~\citep{li2021prefixtuning}.
Such success has also  been shown by the state-of-the-art P-tuning v2 model~\citep{ptuningv2}, which yields performance comparable to fine-tuning on various natural language understanding tasks. 
In addition, Low-Rank Adaptation (LoRA) \citep{hu2021lora} has also emerged as an alternative approach, whereby the weights of rank decomposition matrices that are injected into each layer are optimized in lieu of the full network weights,  reducing trainable parameters while also achieving improvements in accuracy over fine-tuning and other PET methods. 


When  training data is scarce in a task, data augmentation (DA) is a widely used strategy that can boost the performance of deep learning models (see ~\citep{FengGWCVMH21_dasurvey} for a survey on data augmentation in NLP). Then a basic question that needs to be answered is how effective is it when the above  PET frameworks are applied in conjunction with data augmentation. 
To this end, we study three common task-agnostic DA methods, EDA \citep{wei-zou-2019-eda}, Back Translation \citep{sennrich-etal-2016-improving_bt}, and Mixup \citep{Guo2019AugmentingDW} with two common language models BERT \citep{devlin-etal-2019-bert} and RoBERTa \citep{Liu2019RoBERTaAR}, trained using P-tuning v2 \citep{ptuningv2} and LoRA \citep{hu2021lora}. We set up our study across five tasks, including 4 from SuperGLUE \citep{wang2019superglue}, in which the sizes of training data are small. 

We show that data augmentation can increase the accuracy of prefix tuning models. However, the performance of each technique varies depending on the dataset and underlying pretrained model, and the effective techniques differ from fine-tuning. Certain methods can even lead to a notable degradation in performance, particularly when using larger models. 
To better understand the above phenomena, we 
visualize sentence representations of data under prefix tuning, a technique where many augmentation methods fail to get good results, to observe whether any limitations are present with how data is represented that are leading to poor performance in certain cases. Through this, we find that these models struggle more with separating the augmented embeddings from different classes. This observation is further supported by our additional experiments showing a lower robustness to heavily altered augmented data compared to fine-tuning, and an analysis of the cosine similarities between the sentence embeddings of augmented sentences and their original counterparts, where we see that prefix tuning produces highly similar embeddings irrespective of alterations. 
We seek to improve the performance when training on augmented data by adding a contrastive loss term ~\citep{ntxentloss} to 
minimize distances between intra-class embeddings
 in P-tuning v2, helping mitigate the above issues with sentence representations, and resulting in some improvements in accuracy. 
We hope this empirical study helps facilitate future work on leveraging data augmentation when training transformer-based models using parameter efficient tuning methods.  

 

\section{Related Work}

A wide range of data augmentation techniques are used in NLP. Some general types of methods include rule-based techniques that use predetermined, rule-based transformations like synonym replacement that are not dependant on any model architecture \citep{FengGWCVMH21_dasurvey}. Example methods include EDA. Model-based techniques utilize language modelling to generate new data through means like seq2seq translation models as in back translation \citep{sennrich-etal-2016-improving_bt}, masked language modelling using transformers on randomly masked inputs \citep{Garg_2020}, or using generative models like GPT-2 as in \citet{AnabyTavor2020DoNH} where a label-conditioned generation model is obtained by fine-tuning GPT-2 \citep{radford2019language}.

Despite widespread use in computer vision, DA is not as commonly used in the training of NLP models, largely due to the discrete nature of language making it difficult to apply perturbations without compromising the meaning, and the inconsistency in performance of different techniques \citep{FengGWCVMH21_dasurvey}. The effectiveness of DA on fine-tuned models has been studied previously. \citep{LongpreWD20} detail how BT and EDA could not generate consistent benefits on classification tasks with transformer-based models. Similarly, \
\citep{okimura-etal-2022-impact} expand this previous study and test the impact of 12 different DA methods and find little benefit when training on datasets with thousands of examples, but that there is some improvement in performance with very limited data (only a few hundred training cases). Our study conducts a similar examination on the effectiveness of data augmentation with pretrained models, however we focus on prefix tuning and LoRA, and we inspect how the properties of prefix tuning under DA differ from fine-tuning and what practical effect this has, particularly in terms of sentence representations.

\section{Data Augmentation with Parameter Efficient Tuning}
\vspace{-1mm}

\paragraph{Prefix Tuning Approach.} 
We first experiment with prefix tuning, and base our approach on the P-tuning v2 \citep{ptuningv2} implementation. By optimizing prefix length, selectively applying reparameterization, and tuning a randomly-initialized classification head on top of the transformer sentence representations, P-tuning v2 can be applied to smaller language models and across many tasks, including classification, with no drop in performance compared to fine-tuning. P-tuning v2 works by adding prompts of length $p$ tokens to all layers of the language model in the form of sequences of continuous prefix tokens. These continuous embeddings, represented by the vector $ H_m = [h_0, ..., h_p]$ where each $h_i\in \mathbb{R}^d$ has the dimensionality of token embeddings $d$ for each layer $m$ in the transformer-based model, are prefixed to the embeddings of each transformer layer $E_m = [E_{CLS}, E_1, ..., E_n]$ where $n$ is the max sequence length. Thus, each layer of the language model can attend over prompts independently. The overall language model parameters remain frozen during training, with only the parameters of the prefixes and linear head for the classification layer getting optimized. The prefixes can optionally be passed into a reparameterization encoder such as an MLP \citep{li2021prefixtuning}, but this is not always effective and so we only selectively apply this in certain cases where we find it improves performance as in \citet{ptuningv2}. 

\paragraph{LoRA.} 
We test LoRA \citep{hu2021lora}, which is a technique that similarly freezes the original transformer weights, but only tunes trainable pairs of rank-decomposition matrices that are inserted into each transformer layer to approximate the weight updates, greatly reducing the number of parameters needed by avoiding updating the full layer weights while not increasing inference costs.  Instead of updating a pretrained weight matrix of a transformer $W_0 \in R^{d \times k}$
with a gradient-based weight update $\Delta W$  for time step t  to get a new weight matrix as in $W_{t} = W_{0} + \Delta W_t $, LoRA does a low-rank decomposition of the update into two matrices $W_{B} \in R^{d \times r}$ and $W_{A} \in R^{r \times k}$ that act as learnable parameters to represent the new update as $W_{t} = W_{B} W_{A} + W_0 $, with $W_0$ being frozen and where $r$ represents the rank of the decomposition. The authors find applying LoRA to the query and value projection matrices in the attention mechanism leads to the most optimal performance. LoRA often outperforms other PET methods, making it worthy of study in this context \cite{he2022towards}.

\paragraph{Data Augmentation Methods.} 
In this study we focus on \textit{task-agnostic} data augmentation, which are methods that are broadly applicable to a wide range of NLP tasks. 
Many popular DA methods exist and of these, we utilize: (1) \textit{Easy Data Augmentation (EDA)} \citep{wei-zou-2019-eda}, a technique where augmented sentences are created by applying a number of rule-based text editing operations; (2) \textit{Back Translation (BT)} \citep{sennrich-etal-2016-improving_bt} where augmented data is generated by translating text into a target language then translating it back into the original language; (3) \textit{Mixup for Sentence Classification} \citep{Guo2019AugmentingDW}, where we take the senMixup approach by generating synthetic data through interpolation over the classification tokens of random pairs of sentences. All of these have shown an ability to improve model accuracy for limited data \citep{wei-zou-2019-eda,Guo2020NonlinearMO, FengGWCVMH21_dasurvey, btfewshotfabbri} while having a low cost of implementation. 





\section{Experimental Setup}
We focus primarily on small datasets with less than a few thousand annotated data points, since limited data presents challenges in learning generalizable models and hence is a common use case for DA. Previous works have shown DA with pretrained language models tends to be largely ineffective for larger datasets \citep{okimura-etal-2022-impact}. 

We test on 
five text classification datasets. Four are from the SuperGLUE benchmark \citep{wang2019superglue}, consisting of RTE \citep{dagan2006pascal, bar2006second, giampiccolo2007third, bentivogli2009fifth}, CB \citep{demarneffe:cb}, COPA \citep{roemmele2011choice}, and WSC \citep{levesque2011winograd}. These are the smallest datasets in the benchmark at only 2500, 250, 400, 554 training samples respectively. They capture a range of difficulty, and include diverse problems like textual entailment, commonsense-reasoning-based coreference resolution, determining cause or effect, and clause commitment. 
To capture performance on a simpler single sentence sentiment analysis task, we also use a downsampled subset of SST-2 \citep{SocherEtAl2013:RNTN}, consisting of 400 training samples and 100 validation sentences. We experiment with BERT-base \citep{devlin-etal-2019-bert} and RoBERTa-large \citep{Liu2019RoBERTaAR}, which have 110 million and 355 million parameters, respectively. Compared to larger models, their size makes them suitable when storage concerns are present. Each model is trained until convergence and we report results at the epoch where the highest validation accuracy is achieved. We provide the detailed training settings for our models in Appendix \ref{sec:appendix}. Below we describe the settings for each of the data augmentation methods: 

\textbf{EDA}: We apply a combination of the synonym replacement, random swap, random insertion, and random deletion operations to a certain percentage of the words in a sentence to generate multiple augmented sentences per original sentence. For the amount of augmented sentences generated, and the percentage of words affected, we follow the recommended usage guidelines laid out by \citet{wei-zou-2019-eda} which are dependant on the size of the training set. 16 augmented sentences per original are generated for the smallest datasets; CB, COPA, WSC, and SST-2 since they only contain a few hundred training samples samples. The percentage of words in the original text affected for each of the 4 operations is 5\% ($\alpha = 0.05$). 8 sentences are generated per original for the RTE training set, with the same percentage of words being affected as the aforementioned datasets.

\textbf{BT}: 
For our implementation, we back-translate the sentences using a series of English to a target language and target language to English translation models for 4 common languages: French, Spanish, German, and Chinese. The specific models we use are the Helsinki NLP Opus translation modules \citep{tiedemann-2020-tatoeba}. These languages are chosen due to the substantial amount of parallel corpora available for training the respective language models, which yields higher quality translations that are better able preserve the semantics of the original sentence while still introducing some diversity.

\textbf{Mixup}: Inspired by Mixup in image classification \citep{zhang2018mixup,Guo2019MixUpAL}, where random pairs of input images and their labels are linearly interpolated to help generated synthetic images, Mixup can be similarly applied over embeddings for text. As in \citet{Guo2019AugmentingDW}, we adapt Mixup for text classification by interpolating on sentence embeddings (senMixup). In a mini batch of size $N$, we take the same number of random pairs of inputs texts, and interpolate over the $d$-dimensional (768 in the case of BERT-base and 1024 for RoBERTa-large) classification ([CLS]) tokens of the final hidden layer of the transformer encoder produced after each of the inputs is fed into transformer. The interpolated token is then passed into a fully connected later to generate the final softmax prediction vector.
The interpolation for a given pair [CLS] token pair $(x^i, x^j)$ with corresponding one hot label vectors $(y^i, y^j)$, for inputs $i$ and $j$ is conducted as follows, 

\begin{equation} \label{eq1}
\tilde{x}^{ij} = \lambda x^i + (1-\lambda)x^j , 
\end{equation}
\begin{equation} \label{eq1.5}
\tilde{y}^{ij} = \lambda y^i + (1-\lambda)y^j , 
\end{equation}

which results in interpolated vectors $\tilde{x}^{ij}$ and $\tilde{y}^{ij}$. Mixup is parameterized by mixing-ratio $\lambda$, which is different for every pair, and is obtained by sampling from the $Beta(\alpha, \alpha)$ distribution with the hyperparameter $\alpha=1.0$, corresponding to a uniform distribution. This was the recommended setting in \citet{Guo2019AugmentingDW} and consequently we use this value as well. Mixup is done in a similar manner for prefix tuning, except we use the pooled version of the classification token (i.e. one that has gone through a linear layer and tanh activation).

\textbf{Procedure for Augmentation}: For simplicity, since most of the datasets use multiple input sentences/words per training sample (e.g. premise and hypothesis), we apply the static DA methods over the primary (first) input, which is the premise for RTE, CB, and COPA, though we use the full sentence for WSC and SST2. 

\begin{table}
\centering
\caption{Accuracy of fine-tuning BERT and RoBERTa under various augmentation methods with the P-tuning and LoRA trained equivalents.} 

\label{main_table}
\resizebox{\linewidth}{!}{%
\begin{tabular}{llrrrrr} 
\hline
                                   & \multicolumn{6}{c}{\textbf{ Dataset }}                                                                                                      \\ 
\hline
\multicolumn{1}{c}{\textbf{Model}} &         & \multicolumn{1}{l}{rte} & \multicolumn{1}{l}{cb} & \multicolumn{1}{l}{copa} & \multicolumn{1}{l}{wsc} & \multicolumn{1}{l}{sst2}  \\ 
\hline\hline
Fine-tuned BERT                    & No Aug. & 73.5                    & 92.2                   & 69.6                     & 65.4                    & \textbf{91.1}             \\
                                   & EDA     & 67.4                    & \textbf{95.3}          & \textbf{71.1}            & 63.5                    & 88.4                      \\
                                   & Mixup   & \textbf{74.2}           & 92.9                   & 68.8                     & \textbf{66.4}           & 89.3                      \\
                                   & BT      & 71.8                    & 93.8                   & 70.5                     & 63.5                    & \textbf{91.1}             \\ 
\hline
P-tuning v2 BERT                   & No Aug. & 69.4                    & 90.6                   & 72.3                     & \textbf{66.4}           & 89.3                      \\
                                   & EDA     & 68.8                    & 90.1                   & 72.0                     & 63.5                    & 88.4                      \\
                                   & Mixup   & 69.1                    & 91.1                   & 72.0                     & 65.4                    & \textbf{90.2}             \\
                                   & BT      & \textbf{71.2}           & \textbf{94.8}          & \textbf{75.0}            & \textbf{66.4}           & 88.4                      \\ 
\hline
LoRA BERT                          & No Aug. & \textbf{74.2}           & 90.6                   & \textbf{72.7}            & 63.5                    & 89.1                      \\
                                   & EDA     & 69.1                    & 90.6                   & 65.2                     & 63.5                    & 86.6                      \\
                                   & Mixup   & 66.0                    & 89.6                   & 68.0                     & 62.5                    & \textbf{89.8}             \\
                                   & BT      & 70.4                    & \textbf{92.2}          & 67.0                     & \textbf{65.6}           & 87.5                      \\ 
\hline
Fine-tuned RoBERTa                 & No Aug. & \textbf{88.2}           & 98.2                   & \textbf{94.6}            & 63.5                    & \textbf{96.4}             \\
                                   & EDA     & 85.0                    & 96.9                   & 90.2                     & 63.5                    & 94.6                      \\
                                   & Mixup   & 85.4                    & \textbf{98.4}          & 93.8                     & 63.5                    & 95.5                      \\
                                   & BT      & 85.4                    & 96.4                   & 90.2                     & 63.5                    & \textbf{96.4}             \\ 
\hline
P-tuning v2 RoBERTa                & No Aug. & \textbf{87.9}           & \textbf{98.4}          & \textbf{90.0}            & 63.5                    & \textbf{94.6}             \\
                                   & EDA     & 84.6                    & 96.9                   & 87.0                     & \textbf{64.4}           & \textbf{94.6}             \\
                                   & Mixup   & 87.2                    & \textbf{98.4}          & 86.0                     & 63.5                    & 92.9                      \\
                                   & BT      & 84.7                    & \textbf{98.4}          & 85.0                     & 63.5                    & \textbf{94.6}             \\ 
\hline
LoRA RoBERTa                       & No Aug. & 86.5                    & \textbf{96.9}          & \textbf{94.6}            & 63.5                    & 95.5                      \\
                                   & EDA     & 84.4                    & 95.3                   & 89.3                     & 63.5                    & 95.5                      \\
                                   & Mixup   & \textbf{88.2}           & 93.8                   & 89.3                     & 63.5                    & \textbf{96.4}             \\
                                   & BT      & 83.5                    & \textbf{96.9}          & 89.3                     & 63.5                    & \textbf{96.4}             \\
\hline
\end{tabular}
}
\end{table}

\begin{figure*}[htb!]
\centering
\includegraphics[width=.99\textwidth]{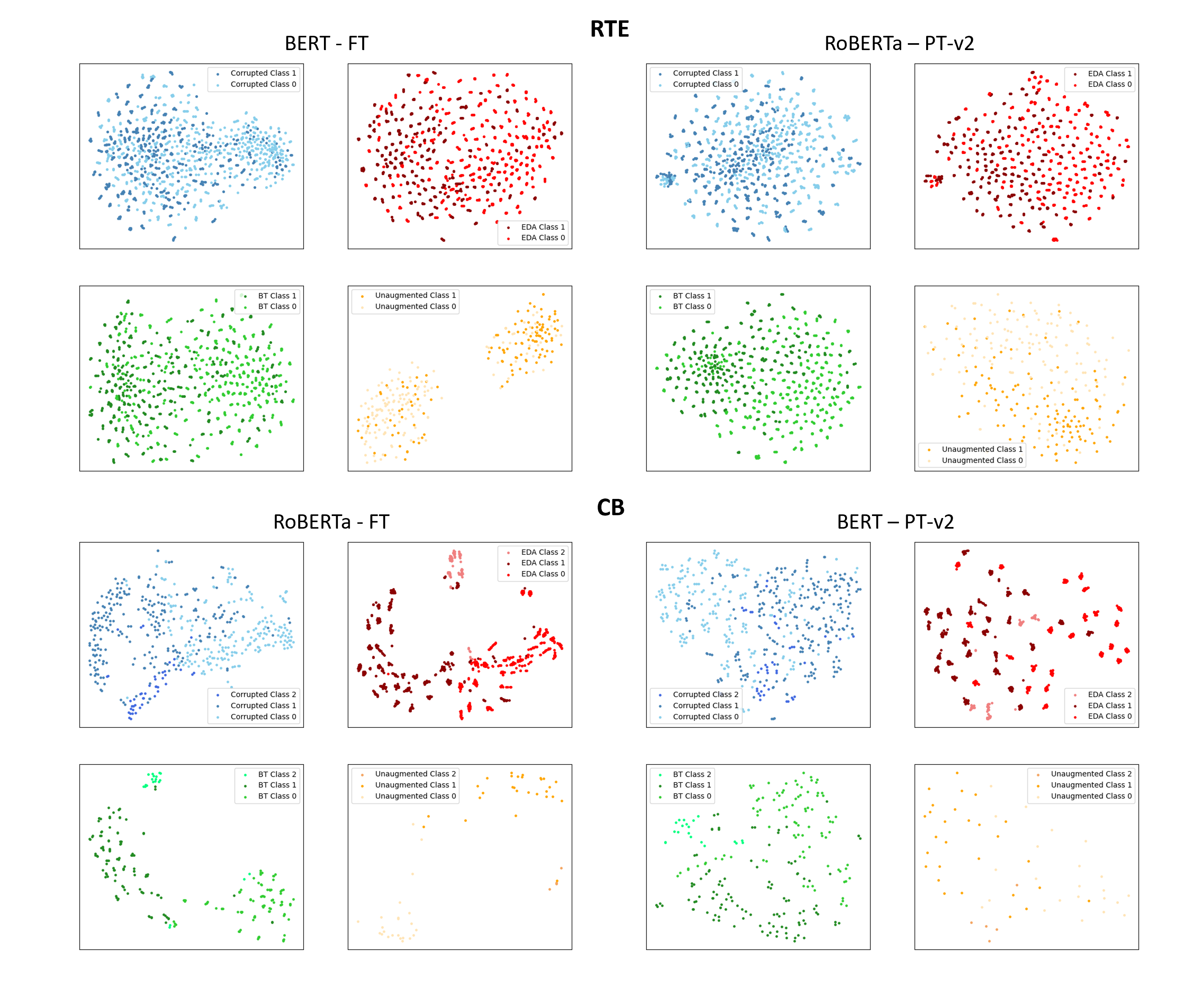}
\caption{The t-SNE latent space for [CLS] token/sentence representation between BERT fine-tuning (top-left), BERT P-tuning v2 (bottom-right) and RoBERTa fine-tuning (bottom-left) and RoBERTa P-tuning v2 (top-right)  on the RTE (top row) and CB (bottom row) datasets. Essentially, for each dataset we compare BERT and RoBERTa using either fine-tuning or P-tuning, while capturing all possible combinations across the two datasets. The resulting representations are obtained from training on clean data and testing on augmented validation data and the original validation data. Within each model scenario, each of the four boxes represents the results of testing that model on a different augmented dataset (blue for corrupted data, red for EDA, green for BT, and orange for the original data). The distribution of each individual class can be seen in these sub-boxes. Across each of the datasets and model types, the sentence representations of data for P-tuning v2 appear more jumbled and less separated across different classes. }
\label{train_aug_tsne}
\end{figure*}

\begin{figure*}[t]
\centering
\includegraphics[width=.7\textwidth]{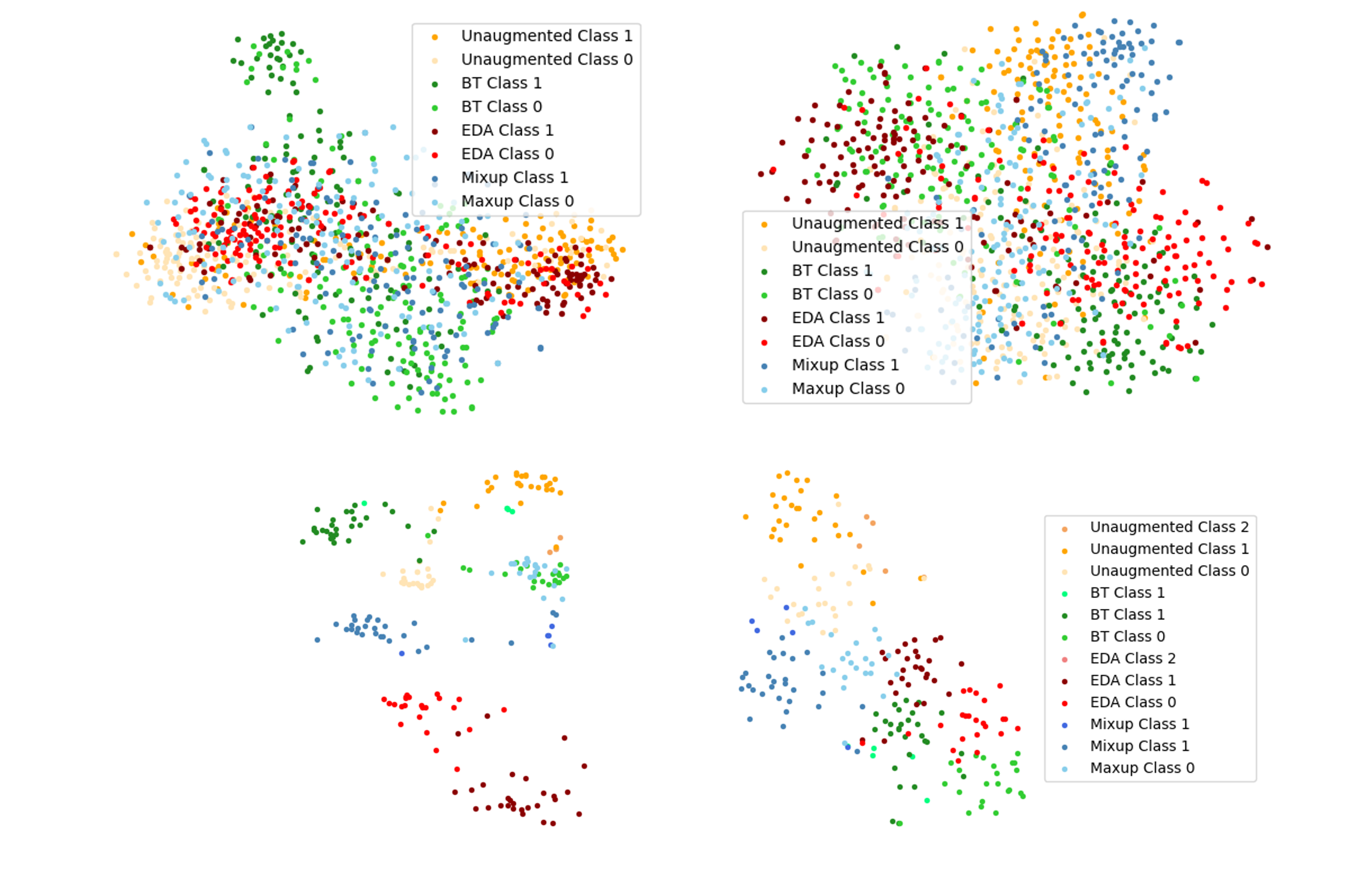}
\caption{The t-SNE latent space for [CLS] token/sentence embeddings between fine-tuning BERT (top-left) and P-tuning RoBERTa (top-right) on RTE, and between fine-tuning (bottom-left) and P-tuning RoBERTa (bottom-right) on CB. The representations are the result of training on augmented data and testing on clean validation data (i.e. legend entries represent which training data was used). }
\label{validate_aug_tsne}
\end{figure*}

\begin{figure}[t]
\centering
\includegraphics[width=.5\textwidth]{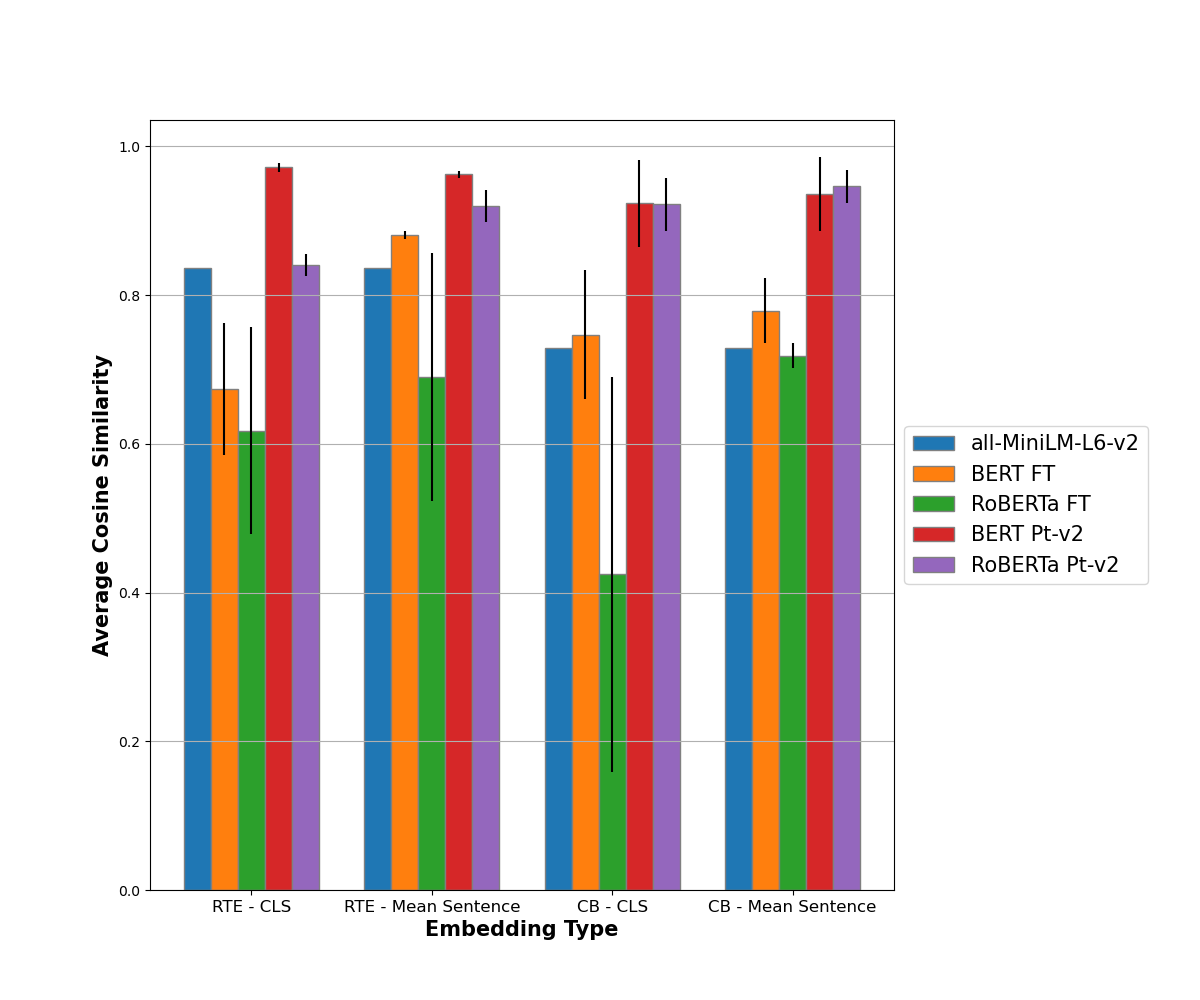}
\caption{Comparison of average cosine similarity between the sentence representations of fine-tuned and P-tuning v2 versions of BERT and RoBERTa across the RTE and CB datasets. We compare the representations of the [CLS] token embeddings and the mean sentence embeddings. We also include a baseline using a Sentence Transformer for an ideal similarity measure. }
\label{sentence_embedding_comparison}
\end{figure}

\section{Experimental Results and Analyses}
\subsection{Data Augmentation with PET}
Table \ref{main_table} displays our primary results. 
The DA methods that achieve the best performance on each dataset and model type vary greatly. Some strong performance gains can be seen when using DA with P-tuning v2, most clearly seen when using BT with BERT on RTE, CB, and COPA. Similar cases exist with LoRA using Mixup on RTE with RoBERTa, and BT on BERT with CB. However, despite the cases of positive performance, frequent degradation to accuracy can be seen in most cases from using DA, particularly with the larger RoBERTa models, and on certain datasets like RTE with P-tuning v2 and COPA with LoRA.
EDA in particular tends to perform poorly when paired with PET methods. WSC sees improvements with EDA, but these models are largely incapable of learning the data properly, so any benefits are likely spurious. 

It is important to take notice of how the best performing method usually differs between fine-tuning, prefix tuning and LoRA, and that the performance of each DA method can change significantly between them as seen with BT on COPA between the the fine-tuned and prefix tuned versions of BERT, hence what works for fine-tuning cannot not be assumed to work for different PET methods. With these results we cannot universally recommend when using DA with prefix tuning or LoRA with limited data, although with careful selection, benefits can still be derived.

\begin{table}
\centering
\caption{Accuracy and average entropy of softmax predictions of P-tuning with fine-tuning when trained on heavily modified training data. P-tuning is generally less robust against the corruptions compared to fine-tuning.  }
\label{corrupt_table}
\resizebox{\linewidth}{!}{%
\begin{tabular}{lrrrrrrrr} 
\hline
\multicolumn{1}{c}{Model} & \multicolumn{2}{l}{rte}                               & \multicolumn{2}{l}{cb}                                & \multicolumn{2}{l}{copa}                              & \multicolumn{2}{l}{sst2}                               \\ 
\hline
                          & \multicolumn{1}{l}{acc} & \multicolumn{1}{l}{ent.} & \multicolumn{1}{l}{acc} & \multicolumn{1}{l}{ent.} & \multicolumn{1}{l}{acc} & \multicolumn{1}{l}{ent.} & \multicolumn{1}{l}{acc} & \multicolumn{1}{l}{ent.}  \\ 
\hline\hline
FT BERT                   & 67.7                    & 11.6                        & 84.4                    & 19.3                        & 65.2                    & 70.2                        & 88.4                    & 59.9                         \\
PT-v2 BERT                & 66.6                    & 63.5                        & 84.4                    & 14.0                        & 73.0                    & 21.4                        & 87.5                    & 39.2                         \\
FT RoBERTa                & 77.6                    & 75.9                        & 90.6                    & 18.9                        & 84.6                    & 26.7                        & 95.5                    & 10.6                         \\
PT-v2 RoBERTa             & 70.6                    & 27.3                        & 89.1                    & 11.4                        & 76.9                    & 19.9                        & 91.1                    & 9.4                          \\
\hline
\end{tabular}
}
\end{table}

\subsection{Embedding Analysis}
\label{tsne_section}
The poor performance of EDA and BT across many scenarios, particularly with P-tuning v2 prompts further study on how prefix tuning learns sentence representations of and under augmented data to identify any potential issues leading to poor performance. To do this, we apply t-SNE \citep{vandermaaten08a} to visualize and understand the latent space sentence embeddings between fine-tuning and P-tuning v2 versions of BERT and RoBERTa across both the RTE and CB datasets where we originally observed stark differences in performance of DA, particularly on RTE since there was a major divide between where DA works well (fine-tuned BERT) versus not well (prefix tuned RoBERTa). We plot the 2-D representation of the [CLS] token in the final hidden layer. Figure \ref{train_aug_tsne} shows the visualizations where we show the representations learned when training on clean data and testing on augmented validation data that was augmented the same way as the training data. We use EDA, BT, and the heavy corruption method described below in Section \ref{pt_data}, along with a baseline using no DA. Through this we can examine how well each method deals with unfamiliar augmented data, and whether the augmented data is represented similarly to the original data. Figure \ref{validate_aug_tsne} displays a different style of representations that are generated from training on augmented data and testing on the clean validation data. This can show how training with DA influences how well the model learns representations between different types of data. In both figures, we observe the same general trends between fine-tuning and prefix tuning, irrespective of model type and dataset. Despite certain models such as P-tuning v2 RoBERTa having stronger predictive performance than their fine-tuned counterparts with BERT, there is a noticeable lack of distinct clustering between class samples of different classes. The sentence representations from different classes of augmented data are less separated. When trained on clean data, the representations of the augmented validation
data are more jumbled and less separate in the representation space as well.
Additionally, when trained using DA, the clustering in the BERT fine-tuned models is more distinct, especially for EDA and BT where P-tuning suffers. Even without augmentation and when using Mixup, the clustering differences remain.
These visualizations demonstrate that there may be difficulties with how prefix tuning is able to process augmented data. Prefix tuning appears to be less robust against augmentation methods like EDA that can change the meaning of the original data, potentially presenting training challenges.

\begin{figure*}
\centering
\includegraphics[width=0.99\textwidth]{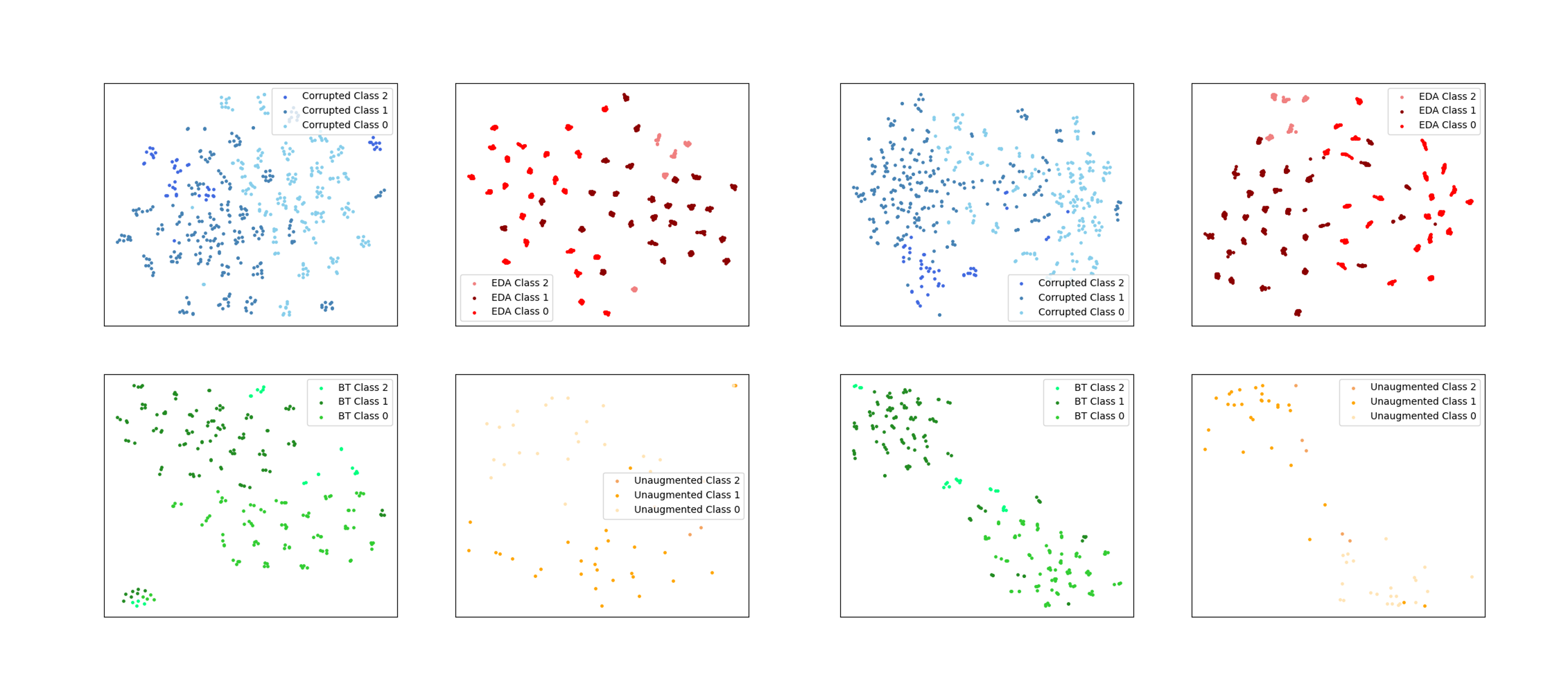}
\caption{t-SNE latent space sentence embedding visualizations between P-tuning v2 RoBERTa with (right) and without (left) the contrastive loss term on the CB dataset. The representations shown are generated from training on EDA augmented data and then testing on various validation data (both augmented and unaugmented) similar to as described in Figure \ref{train_aug_tsne}.}
\label{contrastive_figure}
\end{figure*}

\subsection{Influence of Heavily Perturbed Data}
\label{pt_data}

In view of the generally poor performance of prefix tuning with EDA, and how it has issues with separating the sentence representations, we run experiments to confirm that prefix tuning struggles more with augmented data that is highly perturbed compared to the original. 
To simulate this heavily altered data scenario, we use all four of the EDA operations to modify up to 50\% of the sentence with each of the operations, and generate eight sentences per original on RTE, CB, COPA, and SST2, and we compare the best validation performance of each training method. A performance degradation can be expected since the model should have a more difficult time generalizing when learning from largely incoherent, noisy data as noted in \cite{wei-zou-2019-eda}. As Table \ref{corrupt_table} shows, prefix tuning is generally unable to cope as well as fine-tuning when trained on this data but is usually more confident. 
A possible explanation for this phenomena is that since the parameters of the language models are frozen, the lower number of trainable parameters may have an easier time overfitting to the noise in the augmented data, which explains the low softmax entropy/high prediction confidence. In effect, these results signal the importance of selecting data augmentation methods that can preserve the meaning and structure of the original training data but still provide lexical diversity.

\subsection{Similarity Analysis}
\label{sim_analysis}
To formalize the qualitative analysis provided in Section \ref{tsne_section} with quantitative results and to bring further insight on our results in Section \ref{pt_data}, we run additional experiments comparing the average cosine similarities between the sentence embeddings produced by prefix tuning and fine-tuning. In particular, we compare the original and heavily corrupted versions of sentences. For this, we train fine-tuning and P-tuning v2 versions of BERT and RoBERTa on the clean data of the RTE and CB datasets, and evaluate on heavily corrupted validation constructed in the same manner as described in  Section \ref{pt_data}. We compute the cosine similarity between the sentence embeddings of the original and altered sentences and average across the whole dataset. We use both the [CLS] sentence embeddings as previously, and the mean sentence representation obtained by averaging all tokens in the final hidden state. To get an idea as to the ideal cosine similarities, we also include the average similarity obtained by using a model specifically tuned to measure sentence similarity; Sentence Transformer \textit{all-MiniLM-L6-v2} \cite{reimers-2019-sentence-bert}. We graph our results, which can be seen in Figure \ref{sentence_embedding_comparison}. We average our results over 3 runs and compute the standard deviation. We discover that prefix tuning outputs extremely high similarities despite the data being highly corrupted when compared to fine-tuning, across both types of representations. In most cases, fine-tuning is much closer to the Sentence Transformer, and hence shows a greatly ability to differentiate between the differences in semantics between the two inputs. This signals that prefix-tuning needs a greater ability to be able to differentiate between different data, which may be key to improving its performance on augmented data.

\begin{table}
\centering
\caption{Accuracy with contrastive loss compared to original when training P-tuning v2 models with the static DA methods. There is usually an improvement when using the new loss (Cont.) compared to the original (OG) models.}
\label{contrastive_table}
\resizebox{\linewidth}{!}{%
\begin{tabular}{llrrrrrrrr} 
\hline
        &     & \multicolumn{1}{l}{rte} & \multicolumn{1}{l}{}      & \multicolumn{1}{l}{cb} & \multicolumn{1}{l}{}      & \multicolumn{1}{l}{copa} & \multicolumn{1}{l}{}      & \multicolumn{1}{l}{sst2} & \multicolumn{1}{l}{}       \\ 
\cline{2-10}
        &     & \multicolumn{1}{l}{OG}  & \multicolumn{1}{l}{Cont.} & \multicolumn{1}{l}{OG} & \multicolumn{1}{l}{Cont.} & \multicolumn{1}{l}{OG}   & \multicolumn{1}{l}{Cont.} & \multicolumn{1}{l}{OG}   & \multicolumn{1}{l}{Cont.}  \\ 
\hline\hline
BERT    & EDA & 68.8                    & 72.0                      & 90.1                   & 90.6                      & 72.0                     & 73.2                      & 88.4                     & 89.8                       \\
        & BT  & 71.2                    & 73.2                      & 94.8                   & 93.8                      & 75.0                     & 75.8                      & 88.4                     & 90.6                       \\ 
\hline
RoBERTa & EDA & 84.6                    & 85.1                      & 96.9                   & 98.4                      & 87.0                     & 86.6                      & 94.6                     & 96.1                       \\
        & BT  & 84.7                    & 85.5                      & 98.4                   & 98.4                      & 85.0                     & 85.7                      & 94.6                     & 96.1                       \\
\hline
\end{tabular}
}
\end{table}

\subsection{Benefits of Contrastive Loss}

Given the results in Section \ref{sim_analysis}, it is natural to ponder whether it is possible to improve the performance of DA with prefix tuning by using a method that can promote more defined clusters and greater dissimilarity between semantically different inputs. We study this by adding a contrastive loss term while training P-tuning v2. We generally adopt a version of NT-Xent loss proposed in \cite{ntxentloss}. We adapt their method to this task by creating positive samples in a batch via sampling pairs of inputs with the same class, and creating negative samples by sampling pairs of differing classes. More specifically, given a mini batch of size $N$, we generate an $i$th positive pair $A_i = ({a_i,\hat{a}_i})$ out of $N$ with class $k$, by randomly sampling two inputs with the same class label and using their [CLS] representations. We generate two negative pairs $B_i = ({a_i,\hat{b}_i})$ and $C_i = ({b_i,\hat{a}_i})$ in the same manner where the two randomly sampled inputs $b$ and $\hat{b}$ do not have the same class label as the positive pair. These three pairs are used in the following equation for the contrastive loss, $\mathcal{L} = - \frac{1}{N} \sum_{i=1}^{N} \log \frac{e^{sim(a_{i}, \hat{a}_{i} )/\tau}}{e^{sim(a_{i}, \hat{a}_{i} )/\tau} + e^{sim(a_{i}, \hat{b}_{i} )/\tau} + e^{sim(\hat{a}_{i}, b_{i} )/\tau}} , $ where $sim()$ is a similarity measure for the representations, for which we use cosine similarity. We find that using a high temperature parameter $\tau$ of 0.9 gives the best results. We also scale this additional loss by a parameter $\lambda_{con}$ to avoid too much bias towards towards this loss, which we vary per model. In addition, we use the cosine similarity generated by  Sentence Transformer \textit{all-MiniLM-L6-v2} between the original and augmented training samples to weight the cross entropy loss values to prioritize learning more semantically similar inputs. 

Table \ref{contrastive_table} shows the results when we applied these additional loss terms for the training of P-tuning v2 BERT and RoBERTa models across RTE, CB, COPA, and SST2 using both EDA and BT. We usually observe a rise in accuracy when adding the contrastive term to training with DA, especially with EDA, and in cases where there is no performance gain, the results still remain similar. This signals how contrastive methods may be key to getting good performance using DA with prefix tuning. We do see that in some cases the increase observed is not enough to increase accuracy beyond the model trained without DA, so in these cases these DA techniques are largely inherently too ineffective for that specific task. 

\subsection{Contrastive Loss Representations}

As in Section \ref{tsne_section}, we qualitatively analyze the representations of the models trained using the contrastive loss when they are trained on augmented data. Figure \ref{contrastive_figure} shows how adding the contrastive loss term not only helps the accuracy of augmenting P-tuning v2 RoBERTa on CB with EDA, but that it learns representations for augmented validation data that have more distinct clustering, especially with BT and the unaugmented validation data compared to only using regular cross entropy loss, showing how the method can help induce prefix tuning to learn more useful representations for augmented data. Our contrastive method is rather simple, and not particularly well suited for supervised learning. We do believe that exploring more suitable loss functions like Center Loss \citep{Wen2016ADF} might show even more promising results, which future work can explore.

\section{Conclusion}

We reveal how when limited training data is available, data augmentation can provide crucial improvements in accuracy when training classification models using PET methods, though the benefits are, however, not universal, and limitations exist when using certain methods. We also demonstrated how prefix tuning may have more difficulty learning sentence representations of augmented data. We further showed that contrastive learning can be a solution to these issues, and suggest that more work be done to find similar methods that can be more generalizable.

\section*{Limitations}

Although we present our results across multiple datasets and models, we have largely adhered to natural language understanding tasks. How effective the augmentation methods are on generation-related tasks warrants further study. 

In addition, another limitation is that we do not study the level of perturbations needed to have a negative effect on prefix tuning. Knowing this can shed light on what particular types of transformations are a problem for prefix tuning so that they can be avoided.
Our hypothesis here is that such perturbation levels are sensitive to many factors during training, such as the difficulty of the task and the characteristics of the data augmentation approaches applied, which could be difficult to be quantified.

\section*{Ethics Statement}
Any sort of technique where synthetic text data is generated in an automated fashion, especially with augmentation techniques like BT and EDA that can add or remove random words or completely rephrase sentences, can alter the original meaning or interpretation of a sentence. This has some potential to introduce additional unexpected biases into the training data, particularly when biased models are used to generate synthetic data (e.g. a racially biased translation model). Even with improved performance these biases can cause models to make unfair judgments, which can become a major concern in safety critical domains like in the legal or medical fields. As a consequence, before any data augmentation technique should be applied to PET techniques, there has to be a consideration of the underlying biases that can be introduced through the methods of producing synthetic data.

\section*{Acknowledgements}
We would like to thank the anonymous reviewers for providing valuable feedback for polishing the paper. This work was supported by the Indigenous and Black Engineering and Technology (IBET) Momentum Fellowship and the NSERC Discovery Grants [RGPIN-2018-06415].

\bibliography{anthology,custom}

\begin{thebibliography}{40}
\expandafter\ifx\csname natexlab\endcsname\relax\def\natexlab#1{#1}\fi

\bibitem[{Anaby-Tavor et~al.(2020)Anaby-Tavor, Carmeli, Goldbraich, Kantor,
  Kour, Shlomov, Tepper, and Zwerdling}]{AnabyTavor2020DoNH}
Ateret Anaby-Tavor, Boaz Carmeli, Esther Goldbraich, Amir Kantor, George Kour,
  Segev Shlomov, N.~Tepper, and Naama Zwerdling. 2020.
\newblock {Do Not Have Enough Data? Deep Learning to the Rescue!}
\newblock In \emph{AAAI}.

\bibitem[{Bar~Haim et~al.(2006)Bar~Haim, Dagan, Dolan, Ferro, Giampiccolo,
  Magnini, and Szpektor}]{bar2006second}
Roy Bar~Haim, Ido Dagan, Bill Dolan, Lisa Ferro, Danilo Giampiccolo, Bernardo
  Magnini, and Idan Szpektor. 2006.
\newblock The second {PASCAL} recognising textual entailment challenge.

\bibitem[{Bengio et~al.(2021)Bengio, Lecun, and Hinton}]{bengioDeepLearning}
Yoshua Bengio, Yann Lecun, and Geoffrey Hinton. 2021.
\newblock \href {https://doi.org/10.1145/3448250} {Deep learning for ai}.
\newblock \emph{Commun. ACM}, 64(7):58–65.

\bibitem[{Bentivogli et~al.(2009)Bentivogli, Dagan, Dang, Giampiccolo, and
  Magnini}]{bentivogli2009fifth}
Luisa Bentivogli, Ido Dagan, Hoa~Trang Dang, Danilo Giampiccolo, and Bernardo
  Magnini. 2009.
\newblock The fifth {PASCAL} recognizing textual entailment challenge.

\bibitem[{Dagan et~al.(2006)Dagan, Glickman, and Magnini}]{dagan2006pascal}
Ido Dagan, Oren Glickman, and Bernardo Magnini. 2006.
\newblock The {PASCAL} recognising textual entailment challenge.
\newblock In \emph{Machine learning challenges. evaluating predictive
  uncertainty, visual object classification, and recognising tectual
  entailment}, pages 177--190.

\bibitem[{De~Marneffe et~al.(2019)De~Marneffe, Simons, and
  Tonhauser}]{demarneffe:cb}
Marie-Catherine De~Marneffe, Mandy Simons, and Judith Tonhauser. 2019.
\newblock {The CommitmentBank}: Investigating projection in naturally occurring
  discourse.

\bibitem[{Devlin et~al.(2019)Devlin, Chang, Lee, and
  Toutanova}]{devlin-etal-2019-bert}
Jacob Devlin, Ming-Wei Chang, Kenton Lee, and Kristina Toutanova. 2019.
\newblock \href {https://doi.org/10.18653/v1/N19-1423} {{BERT}: Pre-training of
  deep bidirectional transformers for language understanding}.
\newblock In \emph{Proceedings of the 2019 Conference of the North {A}merican
  Chapter of the Association for Computational Linguistics: Human Language
  Technologies, Volume 1 (Long and Short Papers)}, pages 4171--4186,
  Minneapolis, Minnesota. Association for Computational Linguistics.

\bibitem[{Fabbri et~al.(2020)Fabbri, Han, Li, Li, Ghazvininejad, Joty, Radev,
  and Mehdad}]{btfewshotfabbri}
Alexander~R. Fabbri, Simeng Han, Haoyuan Li, Haoran Li, Marjan Ghazvininejad,
  Shafiq Joty, Dragomir Radev, and Yashar Mehdad. 2020.
\newblock \href {https://doi.org/10.48550/ARXIV.2010.12836} {{Improving Zero
  and Few-Shot Abstractive Summarization with Intermediate Fine-tuning and Data
  Augmentation}}.

\bibitem[{Feng et~al.(2021)Feng, Gangal, Wei, Chandar, Vosoughi, Mitamura, and
  Hovy}]{FengGWCVMH21_dasurvey}
Steven~Y. Feng, Varun Gangal, Jason Wei, Sarath Chandar, Soroush Vosoughi,
  Teruko Mitamura, and Eduard~H. Hovy. 2021.
\newblock \href {https://doi.org/10.18653/v1/2021.findings-acl.84} {{A Survey
  of Data Augmentation Approaches for NLP}}.
\newblock In \emph{ACL/IJCNLP (Findings)}, pages 968--988.

\bibitem[{Garg and Ramakrishnan(2020)}]{Garg_2020}
Siddhant Garg and Goutham Ramakrishnan. 2020.
\newblock \href {https://doi.org/10.18653/v1/2020.emnlp-main.498} {{BAE}:
  {BERT}-based adversarial examples for text classification}.
\newblock In \emph{Proceedings of the 2020 Conference on Empirical Methods in
  Natural Language Processing ({EMNLP})}. Association for Computational
  Linguistics.

\bibitem[{Giampiccolo et~al.(2007)Giampiccolo, Magnini, Dagan, and
  Dolan}]{giampiccolo2007third}
Danilo Giampiccolo, Bernardo Magnini, Ido Dagan, and Bill Dolan. 2007.
\newblock The third {PASCAL} recognizing textual entailment challenge.
\newblock In \emph{Proceedings of the ACL-PASCAL workshop on textual entailment
  and paraphrasing}, pages 1--9. Association for Computational Linguistics.

\bibitem[{Guo(2020)}]{Guo2020NonlinearMO}
Hongyu Guo. 2020.
\newblock {Nonlinear Mixup: Out-Of-Manifold Data Augmentation for Text
  Classification}.
\newblock In \emph{AAAI}.

\bibitem[{Guo et~al.(2019{\natexlab{a}})Guo, Mao, and
  Zhang}]{Guo2019AugmentingDW}
Hongyu Guo, Yongyi Mao, and Richong Zhang. 2019{\natexlab{a}}.
\newblock {Augmenting Data with Mixup for Sentence Classification: An Empirical
  Study}.
\newblock \emph{ArXiv}, abs/1905.08941.

\bibitem[{Guo et~al.(2019{\natexlab{b}})Guo, Mao, and Zhang}]{Guo2019MixUpAL}
Hongyu Guo, Yongyi Mao, and Richong Zhang. 2019{\natexlab{b}}.
\newblock {MixUp as Locally Linear Out-Of-Manifold Regularization}.
\newblock In \emph{AAAI}.

\bibitem[{He et~al.(2022)He, Zhou, Ma, Berg-Kirkpatrick, and
  Neubig}]{he2022towards}
Junxian He, Chunting Zhou, Xuezhe Ma, Taylor Berg-Kirkpatrick, and Graham
  Neubig. 2022.
\newblock \href {https://openreview.net/forum?id=0RDcd5Axok} {{Towards a
  Unified View of Parameter-Efficient Transfer Learning}}.
\newblock In \emph{International Conference on Learning Representations}.

\bibitem[{Hu et~al.(2021)Hu, Shen, Wallis, Allen-Zhu, Li, Wang, and
  Chen}]{hu2021lora}
Edward Hu, Yelong Shen, Phil Wallis, Zeyuan Allen-Zhu, Yuanzhi Li, Lu~Wang, and
  Weizhu Chen. 2021.
\newblock \href {http://arxiv.org/abs/2106.09685} {{LoRA: Low-Rank Adaptation
  of Large Language Models}}.

\bibitem[{Levesque et~al.(2011)Levesque, Davis, and
  Morgenstern}]{levesque2011winograd}
Hector~J Levesque, Ernest Davis, and Leora Morgenstern. 2011.
\newblock The {W}inograd schema challenge.
\newblock In \emph{{AAAI} Spring Symposium: Logical Formalizations of
  Commonsense Reasoning}, volume~46, page~47.

\bibitem[{Li and Liang(2021)}]{li2021prefixtuning}
Xiang~Lisa Li and Percy Liang. 2021.
\newblock \href {http://arxiv.org/abs/2101.00190} {{Prefix-Tuning: Optimizing
  Continuous Prompts for Generation}}.

\bibitem[{Liu et~al.(2021)Liu, Ji, Fu, Du, Yang, and Tang}]{ptuningv2}
Xiao Liu, Kaixuan Ji, Yicheng Fu, Zhengxiao Du, Zhilin Yang, and Jie Tang.
  2021.
\newblock \href {http://arxiv.org/abs/2110.07602} {{P-Tuning v2: Prompt Tuning
  Can Be Comparable to Fine-tuning Universally Across Scales and Tasks}}.
\newblock \emph{CoRR}, abs/2110.07602.

\bibitem[{Liu et~al.(2019)Liu, Ott, Goyal, Du, Joshi, Chen, Levy, Lewis,
  Zettlemoyer, and Stoyanov}]{Liu2019RoBERTaAR}
Yinhan Liu, Myle Ott, Naman Goyal, Jingfei Du, Mandar Joshi, Danqi Chen, Omer
  Levy, Mike Lewis, Luke Zettlemoyer, and Veselin Stoyanov. 2019.
\newblock \href {https://doi.org/10.48550/ARXIV.1907.11692} {{RoBERTa: A
  Robustly Optimized BERT Pretraining Approach}}.

\bibitem[{Longpre et~al.(2020)Longpre, Wang, and DuBois}]{LongpreWD20}
Shayne Longpre, Yu~Wang, and Chris DuBois. 2020.
\newblock \href {https://doi.org/10.18653/v1/2020.findings-emnlp.394} {{How
  Effective is Task-Agnostic Data Augmentation for Pretrained Transformers?}}
\newblock In \emph{EMNLP (Findings)}, pages 4401--4411.

\bibitem[{Loshchilov and Hutter(2019)}]{loshchilov2018decoupled}
Ilya Loshchilov and Frank Hutter. 2019.
\newblock \href {https://openreview.net/forum?id=Bkg6RiCqY7} {{Decoupled Weight
  Decay Regularization}}.
\newblock In \emph{International Conference on Learning Representations}.

\bibitem[{Mosbach et~al.(2021)Mosbach, Andriushchenko, and
  Klakow}]{mosbach2021on}
Marius Mosbach, Maksym Andriushchenko, and Dietrich Klakow. 2021.
\newblock \href {https://openreview.net/forum?id=nzpLWnVAyah} {On the stability
  of fine-tuning {\{}bert{\}}: Misconceptions, explanations, and strong
  baselines}.
\newblock In \emph{International Conference on Learning Representations}.

\bibitem[{Okimura et~al.(2022)Okimura, Reid, Kawano, and
  Matsuo}]{okimura-etal-2022-impact}
Itsuki Okimura, Machel Reid, Makoto Kawano, and Yutaka Matsuo. 2022.
\newblock \href {https://doi.org/10.18653/v1/2022.insights-1.12} {On the impact
  of data augmentation on downstream performance in natural language
  processing}.
\newblock In \emph{Proceedings of the Third Workshop on Insights from Negative
  Results in NLP}, pages 88--93, Dublin, Ireland. Association for Computational
  Linguistics.

\bibitem[{Oord et~al.(2018)Oord, Li, and Vinyals}]{ntxentloss}
Aaron van~den Oord, Yazhe Li, and Oriol Vinyals. 2018.
\newblock \href {https://doi.org/10.48550/ARXIV.1807.03748} {{Representation
  Learning with Contrastive Predictive Coding}}.

\bibitem[{Pfeiffer et~al.(2020)Pfeiffer, R{\"u}ckl{\'e}, Poth, Kamath,
  Vuli{\'c}, Ruder, Cho, and Gurevych}]{pfeiffer2020AdapterHub}
Jonas Pfeiffer, Andreas R{\"u}ckl{\'e}, Clifton Poth, Aishwarya Kamath, Ivan
  Vuli{\'c}, Sebastian Ruder, Kyunghyun Cho, and Iryna Gurevych. 2020.
\newblock {AdapterHub: A Framework for Adapting Transformers}.
\newblock In \emph{Proceedings of the 2020 Conference on Empirical Methods in
  Natural Language Processing: System Demonstrations}, pages 46--54.

\bibitem[{Qin and Eisner(2021)}]{https://doi.org/10.48550/arxiv.2104.06599}
Guanghui Qin and Jason Eisner. 2021.
\newblock \href {https://doi.org/10.48550/ARXIV.2104.06599} {{Learning How to
  Ask: Querying LMs with Mixtures of Soft Prompts}}.

\bibitem[{Radford et~al.(2019)Radford, Wu, Child, Luan, Amodei, and
  Sutskever}]{radford2019language}
Alec Radford, Jeff Wu, Rewon Child, David Luan, Dario Amodei, and Ilya
  Sutskever. 2019.
\newblock {Language Models are Unsupervised Multitask Learners}.

\bibitem[{Reimers and Gurevych(2019)}]{reimers-2019-sentence-bert}
Nils Reimers and Iryna Gurevych. 2019.
\newblock \href {https://arxiv.org/abs/1908.10084} {{Sentence-BERT: Sentence
  Embeddings using Siamese BERT-Networks}}.
\newblock In \emph{Proceedings of the 2019 Conference on Empirical Methods in
  Natural Language Processing}. Association for Computational Linguistics.

\bibitem[{Roemmele et~al.(2011)Roemmele, Bejan, and
  Gordon}]{roemmele2011choice}
Melissa Roemmele, Cosmin~Adrian Bejan, and Andrew~S. Gordon. 2011.
\newblock Choice of plausible alternatives: An evaluation of commonsense causal
  reasoning.
\newblock In \emph{2011 AAAI Spring Symposium Series}.

\bibitem[{Sennrich et~al.(2016)Sennrich, Haddow, and
  Birch}]{sennrich-etal-2016-improving_bt}
Rico Sennrich, Barry Haddow, and Alexandra Birch. 2016.
\newblock \href {https://doi.org/10.18653/v1/P16-1009} {{Improving Neural
  Machine Translation Models with Monolingual Data}}.
\newblock In \emph{Proceedings of the 54th Annual Meeting of the Association
  for Computational Linguistics (Volume 1: Long Papers)}, pages 86--96, Berlin,
  Germany. Association for Computational Linguistics.

\bibitem[{Socher et~al.(2013)Socher, Perelygin, Wu, Chuang, Manning, Ng, and
  Potts}]{SocherEtAl2013:RNTN}
Richard Socher, Alex Perelygin, Jean Wu, Jason Chuang, Christopher Manning,
  Andrew Ng, and Christopher Potts. 2013.
\newblock {Parsing With Compositional Vector Grammars}.
\newblock In \emph{{EMNLP}}.

\bibitem[{Tiedemann(2020)}]{tiedemann-2020-tatoeba}
J{\"o}rg Tiedemann. 2020.
\newblock \href {https://aclanthology.org/2020.wmt-1.139} {The tatoeba
  translation challenge {--} realistic data sets for low resource and
  multilingual {MT}}.
\newblock In \emph{Proceedings of the Fifth Conference on Machine Translation},
  pages 1174--1182, Online. Association for Computational Linguistics.

\bibitem[{van~der Maaten and Hinton(2008)}]{vandermaaten08a}
Laurens van~der Maaten and Geoffrey Hinton. 2008.
\newblock \href {http://jmlr.org/papers/v9/vandermaaten08a.html} {{Visualizing
  Data using t-SNE}}.
\newblock \emph{Journal of Machine Learning Research}, 9(86):2579--2605.

\bibitem[{Wang et~al.(2019)Wang, Pruksachatkun, Nangia, Singh, Michael, Hill,
  Levy, and Bowman}]{wang2019superglue}
Alex Wang, Yada Pruksachatkun, Nikita Nangia, Amanpreet Singh, Julian Michael,
  Felix Hill, Omer Levy, and Samuel~R. Bowman. 2019.
\newblock Super{GLUE}: A stickier benchmark for general-purpose language
  understanding systems.
\newblock \emph{arXiv preprint 1905.00537}.

\bibitem[{Wei and Zou(2019)}]{wei-zou-2019-eda}
Jason Wei and Kai Zou. 2019.
\newblock \href {https://doi.org/10.18653/v1/D19-1670} {{EDA}: Easy data
  augmentation techniques for boosting performance on text classification
  tasks}.
\newblock In \emph{Proceedings of the 2019 Conference on Empirical Methods in
  Natural Language Processing and the 9th International Joint Conference on
  Natural Language Processing (EMNLP-IJCNLP)}, pages 6382--6388, Hong Kong,
  China. Association for Computational Linguistics.

\bibitem[{Wen et~al.(2016)Wen, Zhang, Li, and Qiao}]{Wen2016ADF}
Yandong Wen, Kaipeng Zhang, Zhifeng Li, and Yu~Qiao. 2016.
\newblock {A Discriminative Feature Learning Approach for Deep Face
  Recognition}.
\newblock In \emph{ECCV}.

\bibitem[{Wolf et~al.(2019)Wolf, Debut, Sanh, Chaumond, Delangue, Moi, Cistac,
  Rault, Louf, Funtowicz, Davison, Shleifer, von Platen, Ma, Jernite, Plu, Xu,
  Scao, Gugger, Drame, Lhoest, and Rush}]{huggingface_transformers}
Thomas Wolf, Lysandre Debut, Victor Sanh, Julien Chaumond, Clement Delangue,
  Anthony Moi, Pierric Cistac, Tim Rault, Rémi Louf, Morgan Funtowicz, Joe
  Davison, Sam Shleifer, Patrick von Platen, Clara Ma, Yacine Jernite, Julien
  Plu, Canwen Xu, Teven~Le Scao, Sylvain Gugger, Mariama Drame, Quentin Lhoest,
  and Alexander~M. Rush. 2019.
\newblock \href {https://doi.org/10.48550/ARXIV.1910.03771} {{HuggingFace's
  Transformers: State-of-the-art Natural Language Processing}}.

\bibitem[{Zaheer et~al.(2020)Zaheer, Guruganesh, Dubey, Ainslie, Alberti,
  Ontanon, Pham, Ravula, Wang, Yang, and Ahmed}]{NEURIPS2020_c8512d14}
Manzil Zaheer, Guru Guruganesh, Kumar~Avinava Dubey, Joshua Ainslie, Chris
  Alberti, Santiago Ontanon, Philip Pham, Anirudh Ravula, Qifan Wang, Li~Yang,
  and Amr Ahmed. 2020.
\newblock {Big Bird: Transformers for Longer Sequences}.
\newblock In \emph{NeurIPS}, pages 17283--17297. Curran Associates, Inc.

\bibitem[{Zhang et~al.(2018)Zhang, Cisse, Dauphin, and
  Lopez-Paz}]{zhang2018mixup}
Hongyi Zhang, Moustapha Cisse, Yann~N. Dauphin, and David Lopez-Paz. 2018.
\newblock \href {https://openreview.net/forum?id=r1Ddp1-Rb} {{mixup: Beyond
  Empirical Risk Minimization}}.
\newblock In \emph{International Conference on Learning Representations}.

\end{thebibliography}
\bibliographystyle{acl_natbib}

\clearpage
\newpage

\appendix




\section{Training Settings and Hyperparameters}
\label{sec:appendix}

In this section, we provide details on the general training settings we use along with the hyperparameters. The ideal hyperparameters for each type of model were tuned very carefully by using random search with additional manual tuning, and by largely following the recommendations in \citet{mosbach2021on} to achieve stable, high performance with fine-tuning and LoRA. Likewise, we generally use similar hyperparameters settings as in \citet{ptuningv2} for training P-tuning v2 on the SuperGLUE datasets, since the authors manage to achieve solid results with their settings, although our models still required additional tuning in most cases as our methods differ from their original scope since we incorporate DA. 

The general settings across all the models and datasets include the tokenizer max sequence length being set to 128 tokens, with the exception of WSC and SST-2 which are capped at 64. We use an AdamW optimizer \citep{loshchilov2018decoupled} for all experiments, with a linear scheduler with warmup. For the BERT models we use a gradient clipping value of 1.0. We generally choose batch sizes between 16 or 32. We train each of the models using NVIDIA RTX-3090 GPUs (24GB) with our implementation for the base models being based on HuggingFace Transformers \citep{huggingface_transformers}. We make use of the Sentence-Transformers library as well \cite{reimers-2019-sentence-bert}.

\paragraph{Fine-tuning Settings:}

The settings specific to the fine-tuned models are as follows: across all fine-tuned models we use a warmup rate of 10\%. For BERT we vary the learning rate from 1e-5 and 5e-5 depending on dataset. CB uses 5e-5 for example, though the ideal for the other datasets can vary with 2e-5 for SST-2. The best batch size varies between 16 and 32, though generally a value of 16 performs better. For RoBERTa we use learning rates of either 1e-5 or 2e-5 and a batch size of 16 to achieve stable performance across all datasets and augmentation methods. The number of training epochs we choose is dependant on when the model begins to converge. For EDA this generally falls within 2-4 epochs for all datasets, while for BT it is 4-8. For Mixup and regular training, training epochs are in the range of 8-12, though RTE generally only needs 5-7 epochs. 

\paragraph{P-tuning v2 Settings:}

The ideal hyperparameters for P-tuning v2 strongly vary between the chosen dataset, augmentation method, and model-type, so to save space we only include the ranges over which tune the hyperparameters. We generally keep learning rates between 5e-3 to 5e-2, with the models trained with augmentation methods usually requiring lower rates from 7e-3 to 1e-2. The unaugmented models work best between 1e-2 to 5e-2. The ideal batch size is primarily 16 although 32 is more effective on datasets like RTE. The prefix sequence lengths are kept short, usually between 4 to 32 tokens. Prefix reparameterization is used across most of the datasets, and performs particularly well on CB and SST-2. The hidden dropout probability is consistently set at 0.1. Similarly, the prefix hidden size is kept consistent at 512. We do not use warmup for the prefix models. The number of training epochs to set generally falls within 10-25 for EDA, 20-40 for BT, and for Mixup and regular training 50-120, with the exact number depending on when the model converges to 100\% training accuracy.  The $\lambda_{con}$ parameter we use when training with contrastive loss is set to be between 0.1 and 0.3, with most models using a value of 0.2.

\paragraph{LoRA settings:} In many cases our settings for training the LoRA variations of the models are similar to those described for fine-tuning, though usually LoRA requires shorter training times. Regarding the LoRA specific parameters, rank $r$ and LoRA reparametrization scaling parameter $ \alpha$ , we mirror the settings for BERT-base and RoBERTa-large in the original paper by \citet{hu2021lora} with $r=8$, $\alpha=8$ and $r=8$, $\alpha=16$ for each model type respectively. We use the adapter-transformers library \cite{pfeiffer2020AdapterHub} to implement LoRA.

\end{document}